\documentclass{article}
\usepackage[margin=1in]{geometry}
\usepackage{multirow, multicol, makecell, booktabs, color, colortbl}
\usepackage{tabularx}
\usepackage{graphicx}
\usepackage[dvipsnames]{xcolor}
\usepackage[flushleft]{threeparttable}
\usepackage{longtable}
\usepackage{afterpage}
\usepackage[utf8]{inputenc}
\usepackage[T1]{fontenc}

\usepackage[numbers,sort&compress]{natbib}
\begin{document}

\title{Assessing the Acceptability of a Humanoid Robot for Alzheimer’s Disease and Related Dementia Care Using an Online Survey
}


\author{Fengpei Yuan$^{1}$, Joel G. Anderson$^{2}$, Tami Wyatt$^{2}$,  Ruth Palan Lopez$^{3}$, \\Monica Crane$^{4}$, Austin Montgomery$^{4}$ and Xiaopeng Zhao$^{1}$%
\thanks{$^{1}$The authors are with the Department of Mechanical, Aerospace and Biomedical Engineering,
        University of Tennessee, Knoxville, TN, USA
        {\tt\small (fyuan6@vols.utk.edu; xzhao9@utk.edu)}}%
\thanks{$^{2}$The authors are with College of Nursing, University of Tennessee, Knoxville, TN, USA
        {\tt\small (jande147@utk.edu; twyatt@utk.edu)}}%
\thanks{$^{3}$Ruth Palan Lopez is with Institute of Health Professions, Boston, MA, USA
        {\tt\small (rlopez@mghihp.edu)}}
\thanks{$^{4}$The authors are with Genesis Neuroscience Clinic, Knoxville, TN, USA
        {\tt\small (mcrane@genesisneuro.com; clinical.coordinator@genesisneuro.com)}}
}

\maketitle

\begin{abstract}
In this work, an online survey was used to understand the acceptability of humanoid robots and users’ needs in using these robots to assist with care among people with Alzheimer’s disease and related dementias (ADRD), their family caregivers, health care professionals, and the general public. From November 12, 2020 to March 13, 2021, a total of $631$ complete responses were collected, including $80$ responses from people with mild cognitive impairment or ADRD, $245$ responses from caregivers and health care professionals, and $306$ responses from the general public. Overall, people with ADRD, caregivers, and the general public showed positive attitudes towards using the robot to assist with care for people with ADRD. The top three functions of robots required by the group of people with ADRD were reminders to take medicine, emergency call service, and helping contact medical services. Additional comments, suggestions, and concerns provided by caregivers and the general public are also discussed.
\end{abstract}

\section{Introduction}
\label{intro}
As the number and proportion of people aged 65 or older continue to increase worldwide, the number of adults living with dementia is also growing. According to the World Alzheimer Report 2018 \cite{Patterson2018world}, there were 50 million people living with dementia in 2018 around the world, with one new case of dementia every 3 seconds. In the U.S., an estimated 6.2 million adults live with Alzheimer’s disease and related dementias (ADRD) \cite{alzheimer20212021}. According to 2021 Alzheimer's Disease Facts and Figures report, more than 11 million unpaid caregivers (e.g., family members, friends and relatives) provided care for people with ADRD in 2020 \cite{alzheimer20212021}. This long-duration, time-intensive care can negatively impact caregivers’ physical and mental health and contribute to increased risk for depression and poor health \cite{brodaty2009family}.
With the advancement of robotics and information and communication technologies (ICTs), social robots have become one of the two most common forms of assistive technology to support people with ADRD \cite{kruse2020evaluating}. As an augmentation of human caregivers with respect to the substantial health care labor shortage and the high burden of caregiving, robots may provide care with high repeatability and without any complaints or fatigue \cite{taheri2015clinical}. Moreover, a meta-analysis by Li \cite{li2015benefit} comparing how people interacted with physical robots and virtual agents showed that physically present robots were found to be more persuasive, perceived more positively, and resulted in better user performance compared to virtual agents. Furthermore, robots can facilitate social interaction, communication, engagement, and positive mood to improve the performance of the system \cite{shibata2012therapeutic}. For example, a recent study \cite{pino2020humanoid} showed that older adults with mild cognitive impairment (MCI) who received memory training through a humanoid social robot (NAO) achieved more visual gaze, fewer symptoms of depression, and better therapeutic behavior. In a study \cite{manca2021impact} to investigate the impact of how humanoid robots support older adults with MCI in serious games, which are games aimed to go beyond mere entertainment, but for helping cognitive assessment, stimulation and rehabilitation \cite{mccallum2013dementia}, the same music-based memory serious game was implemented using a robot (i.e., Pepper) and a tablet. Older adults with MCI reported higher engagement in the games with the robot than those using the tablet. The results also suggested the robot produced more empathetic and emotional responses from the participants than the tablet.

Using the principle of user-centered design during the development of robot as a complement to care people with ADRD, it is imperative to understand users’ requirements before design and implementation of a specific protocol of a robot system \cite{hartson2018ux,di2018multi}. As found in a literature review of factors affecting the acceptability of social robots among older adults (including people with ADRD and MCI), acceptability is likely to be improved if the robot meets users’ emotional, psychological, social, and environmental needs \cite{whelan2018factors}. Surveys are a popular, practical approach to learn the acceptability and perceptions of robots and users’ needs and requirements. For example, Sancarlo et al. \cite{sancarlo2016mario} conducted a multicenter survey in Italy and Ireland about companion robots to understand the needs of caregivers as well as their preferences that would improve their ability in assisting those with ADRD. A majority ($64.5\%$) of caregivers said that the robot MARIO would be useful or moderately useful to improve quality of life, quality of care, safety, and emergency communications. Cavallo et al. \cite{cavallo2018robotic} conducted a user satisfaction and acceptability study regarding use of the Robot-Era system to provide robotic services to permit older people to remain in their homes independently. They recruited 45 older users who had normal cognitive functioning and a minimum required autonomy in performing daily activities. Participants’ feedback showed that the Robot-Era system has the potential to be developed as a socially acceptable and believable provider of robotic services to facilitate older people to live independently in their homes.

However, attitudes towards using robot to assist dementia care can be different across cultures and countries \cite{coco2018care,suwa2020exploring}. For example, in a cross-sectional study by Coco et al. \cite{coco2018care} in Finland and Japan, Japanese care personnel overall showed more positive assessment of the usefulness of robots than their Finnish counterparts, while more Finnish care personnel showed fear related to the introduction of care robots. Moreover, users' experience with technology is also a significant moderating factor for the acceptance of socially assistive robots by older adults \cite{flandorfer2012population}. As the presence and use of smartphones, computers, and Internet have grown considerably over the past few decades \cite{ryan2018computer}, we would like to know people's recent opinions towards towards using robots to supplement caring people with ADRD. Furthermore, feedback from not only from the primary end users (i.e., people with MCI or ADRD) but also from other stakeholders such as family caregivers, paid in-home caregivers, therapists, nurses, and other health care providers will be useful in the design process. In addition, considering that the general public (e.g., young age group) could possibly be the stakeholders in the future, we also took into consideration of their feedback. We used an online, anonymous survey to collect responses from these users.

We created the survey based on the Almere model \cite{heerink2010assessing} and questionnaires adopted from previous relevant studies \cite{cavallo2018robotic,sancarlo2016mario}. There are several goals in this survey study: 1) to understand public attitudes and acceptance of using a humanoid social robot, Pepper (SoftBank Robotics) \cite{pandey2018mass}, to assist in caring for people with ADRD; 2) to explores users’ needs and requirements for a robot for ADRD care.

\section{Methods}
\label{methods}
\subsection{Participants}
\label{participants}
The study protocol was approved by the Institutional Review Board (IRB) of the University of Tennessee, Knoxville. Adults over the age of 18 were eligible to participate. There were no specific exclusion criteria. Consent was implied by completing the survey. The survey was distributed via social media (i.e., Facebook, Twitter, LinkedIn, and WeChat), email, and listservs.

\subsection{Instrument}
A structured, anonymous questionnaire written in English was created using the web survey design tool QuestionPro. At the beginning of the survey, there was an approximately three-minute video in which a humanoid social robot Pepper \cite{pandey2018mass}, named Tammy Pepper Knox or Tammy, introduces how a robot could help care for someone with ADRD. After watching the video, participants were asked to four questions regarding demographic information, including age, gender, level of education, and familiarity with technology (i.e., smartphones, tablets, computers and robots). A branching question, "Are you currently diagnosed with any of the following conditions?" was used to identify those with MCI and ADRD. Considering the cognitive and physical limitations of people with ADRD, there are only 11 questions/items for those indicating a diagnosis of MCI or ADRD, and 32 questions for other participants. For those participants with ADRD, 11 statements regarding their opinions of the robot’s appearance, voice, height, and intention to use the robot was provided using Likert-type scale ratings. A second branching question, “Have you ever provided Alzheimer’s care as any of the following types of caregivers?”, was used to identify caregivers, health care professionals, and the general public. Those who identified as caregivers were asked how many hours they provided each week: less than four hours, four to eight hours, or more than eight hours. Caregiver and other participants were asked 32 questions related to acceptability, function, impact, and ethical concerns of using a robot Tammy for ADRD care. Participants could skip any question they did not want to answer. The survey for caregivers and the general public ended with an open-ended question to provide additional comments, suggestions, or concerns regarding the development of a robot to assist in providing ADRD care.

In addition to the aforementioned items, participants' physical locations (e.g., country and region) were automatically collected and stored by the web survey tool, QuestionPro, through their IP addresses \cite{Questionpro2021geo}. 

\subsection{Data analysis}
For each question, we calculated basic descriptive statistics as appropriate, e.g., mean and standard deviation (SD), frequency and percentage. In the case of missing response to a question, the missing data were excluded. Then, considering that the data are not normally distributed, the Mann-Whitney U test was used to compare the caregiver group versus the other public group and female versus male within the same group. Furthermore, we investigated the relationships of responses among people with ADRD associated with perceptions/acceptability of the robot by calculating the Spearman’s correlation. During computing pairwise correlation of items, the missing data were excluded.

\section{Results}
From November 12, 2020 to March 13, 2021, a total of 2518 people viewed the survey. Of those, 944 people started the survey and 631 people completed it, with an average time of 11 minutes. The survey was considered completed when a respondent reached the Thank You page, although leaving some questions unanswered (i.e., skipping some questions). Among all the complete responses, $601$ $(95.25\%)$ were from U.S. and $30$ $(4.75\%)$ from 12 other countries (e.g., China, Germany, Japan, Australia, and Slovakia). The detailed response distribution by country is listed in Table \ref{table:CountryDistribution}. Within the U.S., responses were collected from participants in 32 states. $485$ $(80.70\%)$ responses were from the state of Tennessee (Table \ref{table:US_StatesDistribution}).
\begin{table}[t]
\caption{Response distribution of participants around the world}
\label{table:CountryDistribution}
\begin{center}
\begin{tabular}{lcc}
\hline\noalign{\smallskip}
{Country N = 13} & Count & Percentage \\
  \noalign{\smallskip}\hline\noalign{\smallskip}
  United States & {$601$} & {$95.25\%$}\\
  China & {$13$} & {$2.06\%$}\\
  Hong Kong & {$4$} & {$0.63\%$}\\
  Germany & {$3$} & {$0.48\%$}\\
  Japan & {$2$} & {$0.32\%$}\\
  Australia & {$1$} & {$0.16\%$}\\
  Canada & {$1$} & {$0.16\%$}\\
  Europe & {$1$} & {$0.16\%$}\\
  United Kingdom & {$1$} & {$0.16\%$}\\
  India & {$1$} & {$0.16\%$}\\
  Saudi Arabia & {$1$} & {$0.16\%$}\\
  Slovakia & {$1$} & {$0.16\%$}\\
  Taiwan & {$1$} & {$0.14\%$}\\
  \noalign{\smallskip}\hline
\end{tabular}
\end{center}
\end{table}
\begin{table}[hbt!]
\caption{Response distribution of participants within US}
\label{table:US_StatesDistribution}
\begin{center}
\begin{tabular}{lcc} 
\hline\noalign{\smallskip}
{States N = 32} & Count & Percentage \\
\noalign{\smallskip}\hline\noalign{\smallskip}
AL &7& $1.16\%$\\
AR & 3 & $0.50\%$\\
CA & 8 & $1.33\%$\\
CO & 3 & $0.50\%$\\
DC & 3 & $0.50\%$\\
FL & 9 & $1.50\%$\\
GA & 6 & $1.00\%$\\
IA & 2 & $0.33\%$\\
IL & 2 & $0.33\%$\\
IN & 2 & $0.33\%$ \\
KY & 5 & $0.83\%$ \\
LA & 2 & $0.33\%$\\
MA & 3 & $0.50\%$\\
MD & 4 & $0.67\%$\\
MI & 3 &$0.50\%$\\
MO & 1 & $0.17\%$\\
NC & 3 & $0.50\%$\\
NJ & 1 & $0.17\%$\\
NM & 1 & $0.17\%$\\
NV & 1 & $0.17\%$\\
NY & 3 & $0.50\%$\\
OH & 2 & $0.33\%$\\
OK & 3 & $0.50\%$\\
OR & 1 & $0.17\%$\\
PA & 2 & $0.33\%$\\
SC & 5 & $0.83\%$\\
TN & 485 & $80.70\%$\\
TX & 3 & $0.50\%$\\
VA & 5 & $0.83\%$\\
WA & 1 & $0.17\%$\\
WI & 1 & $0.17\%$\\
WV & 1 & $0.17\%$\\
Unknown & 20 & $3.31\%$\\
\noalign{\smallskip}\hline
\end{tabular}
\end{center}
\end{table}

\subsection{Demographic information}
Among the 631 complete responses, 80 were from people with MCI or ADRD or other dementia, 245 from caregivers, and 306 from the general public. On average, it took those with ADRD 10.04 minutes to complete the survey, 17.57 minutes for caregivers, and 16.61 minutes for the general public. The relative number and percentage of participants in each group in terms of gender, age range, highest level of education, and familiarity with technology are listed in Table \ref{Table:DemographicInfo}.
\begin{table*}[htp]
\caption{Demographic information of participants}
\label{Table:DemographicInfo}
\begin{center}
\begin{tabular}{llll} 
\hline\noalign{\smallskip}
{Demographic variables} & \makecell[c]{People with MCI\\or ADRD $n=80$} & \makecell[c]{Caregivers\\$n=245$} & \makecell[c]{General public\\$n=306$} \\
  \noalign{\smallskip}\hline\noalign{\smallskip}
  {Gender (n,$\%$)} & & & \\
  {\hspace{0.5cm}Female} & {$44$ $(55.00\%)$} & {$184$ $(75.10\%)$} & {$177$ $(57.84\%)$}\\
  \hspace{0.5cm}Male & {$35$ $(43.75\%)$} & {$58$ $(23.67\%)$} & {$124$ $(40.52\%)$} \\
  \hspace{0.5cm}{Prefer not to answer} & {$1$ $(1.25\%)$}  & {0} & {$1$ $(0.33\%)$}\\
  \hspace{0.5cm}{Other} & {0}  & {0} & {$1$ $(0.33\%)$}\\
  {Age (years) (n,$\%$)} & & & \\
  \hspace{0.5cm}Under 50 & {$8$ $(10.00\%)$}  & {$54$ $(22.04\%)$} & {$104$ $(33.99\%)$}\\
  \hspace{0.5cm}50-65 & {$17$ $(21.25\%)$}  & {$86$ $(35.10\%)$} & {$52$ $(16.99\%)$} \\
  \hspace{0.5cm}65-75 & {$27$ $(33.75\%)$}  & {$76$ $(31.02\%)$} & {$94$ $(30.72\%)$}\\
  \hspace{0.5cm}75-85 & {$22$ $(27.50\%)$}  & {$27$ $(11.02\%)$} & {$51$ $(16.67\%)$}\\
  \hspace{0.5cm}Above 85 & {$6$ $(7.50\%)$} & {$2$ $(0.82\%)$} & {$5$ $(1.63\%)$}\\
  {Highest level of education (n,$\%$)} & & & \\
  \hspace{0.5cm}6th grade or less & {$0$} & {$1$ $(0.41\%)$} & {$0$}\\
  \hspace{0.5cm}7th-11th grade & {$2$ $(2.50\%)$}  & {$0$} & {$2$ $(0.65\%)$}\\
  \hspace{0.5cm}High school graduate & {$8$ $(10.00\%)$} & {$10$ $(4.08\%)$} & {$18$ $(5.88\%)$}\\
  \hspace{0.5cm}Some college & {$20$ $(25.00\%)$}  & {$53$ $(21.63\%)$} & {$54$ $(17.65\%)$}\\
  \hspace{0.5cm}College graduate & {$17$ $(21.25\%)$}  & {$61$ $(24.90\%)$} & {$94$ $(30.72\%)$}\\
  \hspace{0.5cm}Post-graduate & {$33$ $(41.25\%)$}  & {$120$ $(48.98\%)$} & {$138$ $(45.10\%)$}\\
  {Familiarity with technology (Mean$\pm$SD)} & & & \\ \hspace{0.5cm}Smartphones & {$3.02\pm1.02$} & {$3.76\pm0.49$} & {$3.63\pm0.64$}\\ 
  \hspace{0.5cm}Tablets & {$2.80\pm1.10$}  & {$3.52\pm0.73$} & {$3.40\pm0.82$}\\
  \hspace{0.5cm}Computers & {$3.09\pm0.99$}  & {$3.74\pm0.49$} & {$3.63\pm0.57$}\\
  \hspace{0.5cm}Robots & {$1.58\pm0.81$} & {$1.91\pm0.91$} & {$1.96\pm0.96$}\\
  {Types of caregivers (Multiple answer; n,$\%$)} & & & \\
  \makecell[l]{\hspace{0.5cm}Informal caregiver (e.g., family,\\\hspace{1cm}relative or friend)} & NA & {$212$ $(86.53\%)$} & NA\\ 
  \hspace{0.5cm}Paid in-home caregiver & NA & {$13$ $(5.31\%)$} & NA\\ 
  \hspace{0.5cm}Nurse & NA & {$36$ $(14.69\%)$} & NA\\
  \hspace{0.5cm}Therapist & NA & {$14$ $(5.71\%)$} & NA\\
  \hspace{0.5cm}Doctor & NA & {$8$ $(3.27\%)$} & NA\\
  \hspace{0.5cm}Psychologist & NA & {$2$ $(0.82\%)$} & NA\\
  {Caregiving hours per week (n,$\%$)} & & & \\
  \hspace{0.5cm}Under 4 hours & NA & {$141$ $(57.55\%)$} & NA\\ 
  \hspace{0.5cm}4-8 hours & NA & {$27$ $(11.02\%)$} & NA\\ 
  \hspace{0.5cm}Above 8 hours & NA & {$72$ $(29.39\%)$} & NA\\
  \noalign{\smallskip}\hline
\end{tabular}
\begin{tablenotes}
            \item \textit{Note:} NA = Not applicable.
            \item Among caregiver respondents, there were {$3$} participants who skipped the question associated with gender and {$5$} participants who skipped the question associated with hours of caregiving.
            \item Among respondents from the general public, there were {$3$} participants who skipped the question associated with gender.
            \item For the items associated with familiarity with technology (i.e., Rows \textit{Smartphones}, \textit{Tablets}, \textit{Computers} and \textit{Robots}), 1=Not at all familiar, 2=Slightly familiar,3=Moderately familiar, 4=Very familiar.
        \end{tablenotes}
\end{center}
\end{table*}

\begin{figure}[ht]
    \centering
    \includegraphics[width=7.2cm]{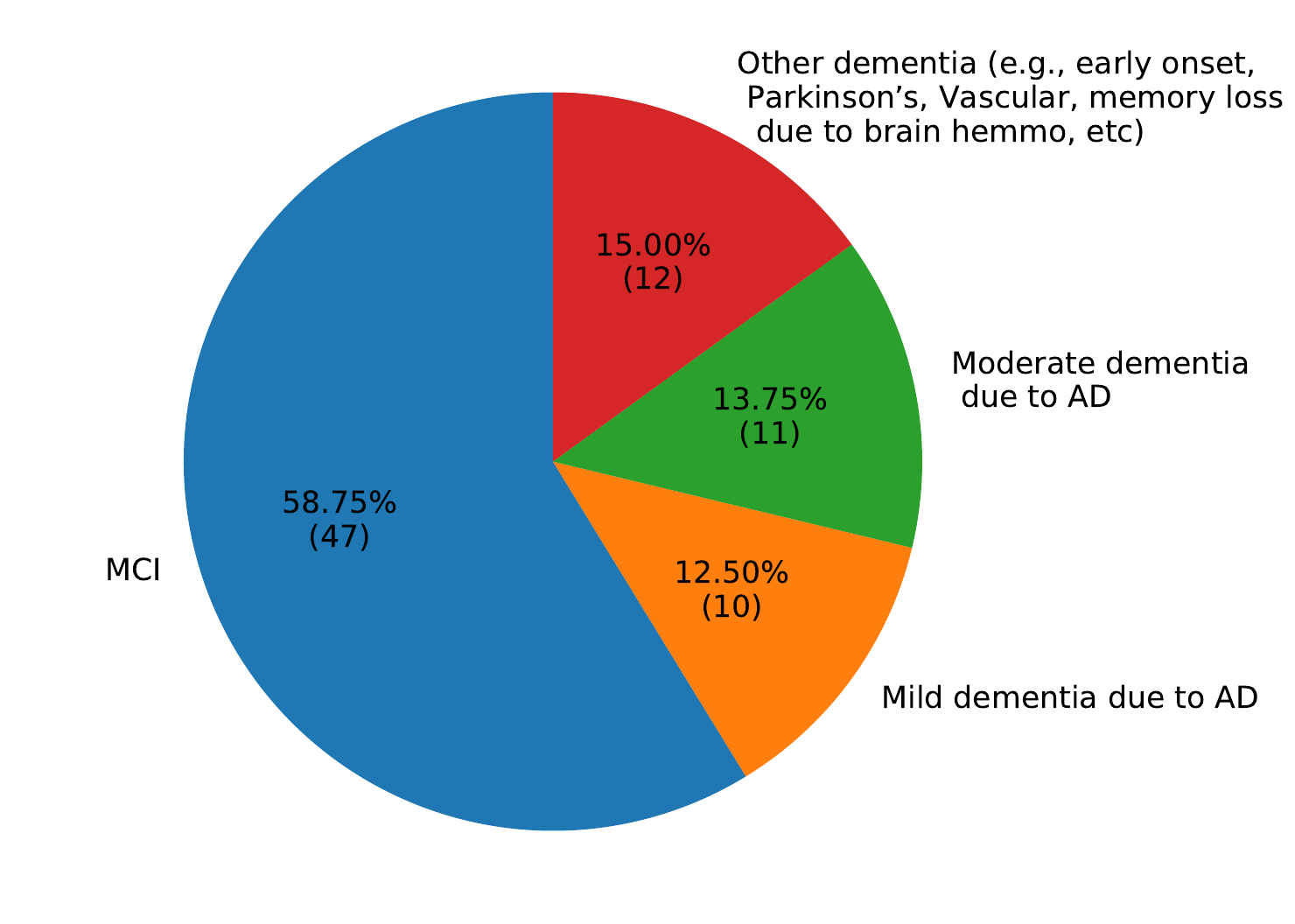}
    \caption{Disease distribution among patient group}
    \label{fig:diseaseDistribution}
\end{figure}

The disease distribution of those with MCI or ADRD is illustrated in Figure \ref{fig:diseaseDistribution}. The majority ($n = 47$) were currently diagnosed with MCI, $10$ with mild dementia due to Alzheimer’s disease, $11$ with moderate dementia due to Alzheimer’s disease. In addition, $12$ participants chose other dementia, including bouts of amnesia (unknown causes), early onset dementia, hardening of arteries, memory loss due to brain hemorrhage, age-related dementia (not diagnosed as Alzheimer’s), either mild or moderate dementia (not sure), Parkinson’s dementia, and vascular dementia.

Participants in the caregiver group provided care as an informal caregiver ($n=212$), paid in-home caregiver ($n=13$), nurse ($n=36$), therapist ($n=14$), physician ($n=8$), and psychologist ($n=36$). The majority ($57.55\%$) of caregivers indicated they provided care for fewer than 4 hours per week, $11.02\%$ provided care for 4–8 hours, and $29.39\%$ for more than 8 hours.

\subsection{Perceptions of robot among people with MCI or ADRD}
\label{subsection_resultPerceptbyPatient}
\begin{center}
\begin{table*}
    \caption{Percentage of responses by people with MCI or ADRD.}
    \begin{tabular}{p{0.8cm}p{1.8cm}p{2.2cm}p{1.8cm}p{1.8cm}p{1.8cm}p{1.8cm}}
    \hline\noalign{\smallskip}
    {Items} & No=1 & Probably not=2 & Maybe=3 & Probably=4 & Yes=5 & {Mean(SD)}\\
    \noalign{\smallskip}\hline\noalign{\smallskip}
    {Q1} & $0$ & $0$ & $11$ $(13.75\%)$ & $13$ $(16.25\%)$ & $56$ $(70.00\%)$ & $4.56$ $(0.73)$\\
    Q2 & $7$ $(8.75\%)$ & $11$ $(13.75\%)$ & $16$ $(20.00\%)$ & $18$ $(22.50\%)$ & $28$ $(35.00\%)$ & $3.61$ $(1.33)$\\
    Q5 & $5$ $(6.25\%)$ & $5$ $(6.25\%)$ & $22$ $(27.50\%)$ & $17$ $(21.25\%)$ & $31$ $(38.75\%)$ & $3.80$ $(1.21)$\\
    Q6 & $3$ $(3.80\%)$ & $4$ $(5.06\%)$ & $13$ $(16.46\%)$ & $34$ $(43.04\%)$ & $25$ $(31.65\%)$ & $3.94$ $(1.02)$\\
    Q7 & $27$ $(33.75\%)$ & $26$ $(32.50\%)$ & $15$ $(18.75\%)$ & $8$ $(10.00\%)$ & $4$ $(5.00\%)$ & $2.20$ $(1.16)$\\
    Q8 & $7$ $(8.86\%)$ & $8$ $(10.13\%)$ & $16$ $(20.25\%)$ & $14$ $(17.72\%)$ & $34$ $(43.04\%)$ & $3.76$ $(1.34)$\\
    Q9 & $2$ $(2.50\%)$ & $4$ $(5.00\%)$ & $18$ $(22.50\%)$ & $28$ $(35.00\%)$ & $28$ $(35.00\%)$ & $3.95$ $(1.01)$ \\
    \noalign{\smallskip}\hline
    \end{tabular}
    \label{Table:PercentFreq_Patient}
    \begin{tablenotes}
            \item \textit{Note:} Please see Supplementary Appendix 1 for Items Q1-Q9.
        \end{tablenotes}
\end{table*}
\end{center}
With respect to robot height (4 feet tall), the majority of people with ADRD ($66$ out of $79$ response, with $1$ participant skipping this question) thought the height was about right, while $11$ and $2$ participants thought it too short or too tall, respectively. In terms of the scary level of robot Tammy, $66$, $12$ and $2$ out of $80$ of those with MCI or ADRD separately found the robot not scary at all, slightly scary and moderately scary. None reported the robot as very scary. The responses (i.e., frequency, percentage, mean and standard deviation) to other questions associated with perception and acceptability of robot are listed in Table \ref{Table:PercentFreq_Patient}. Except for one item (i.e., "If you use the robot Tammy, would you be afraid to break something of the robot?"), the mean values of responses to items were greater than $3$.

\begin{figure}[h]
    \centering
    \includegraphics[width=8cm]{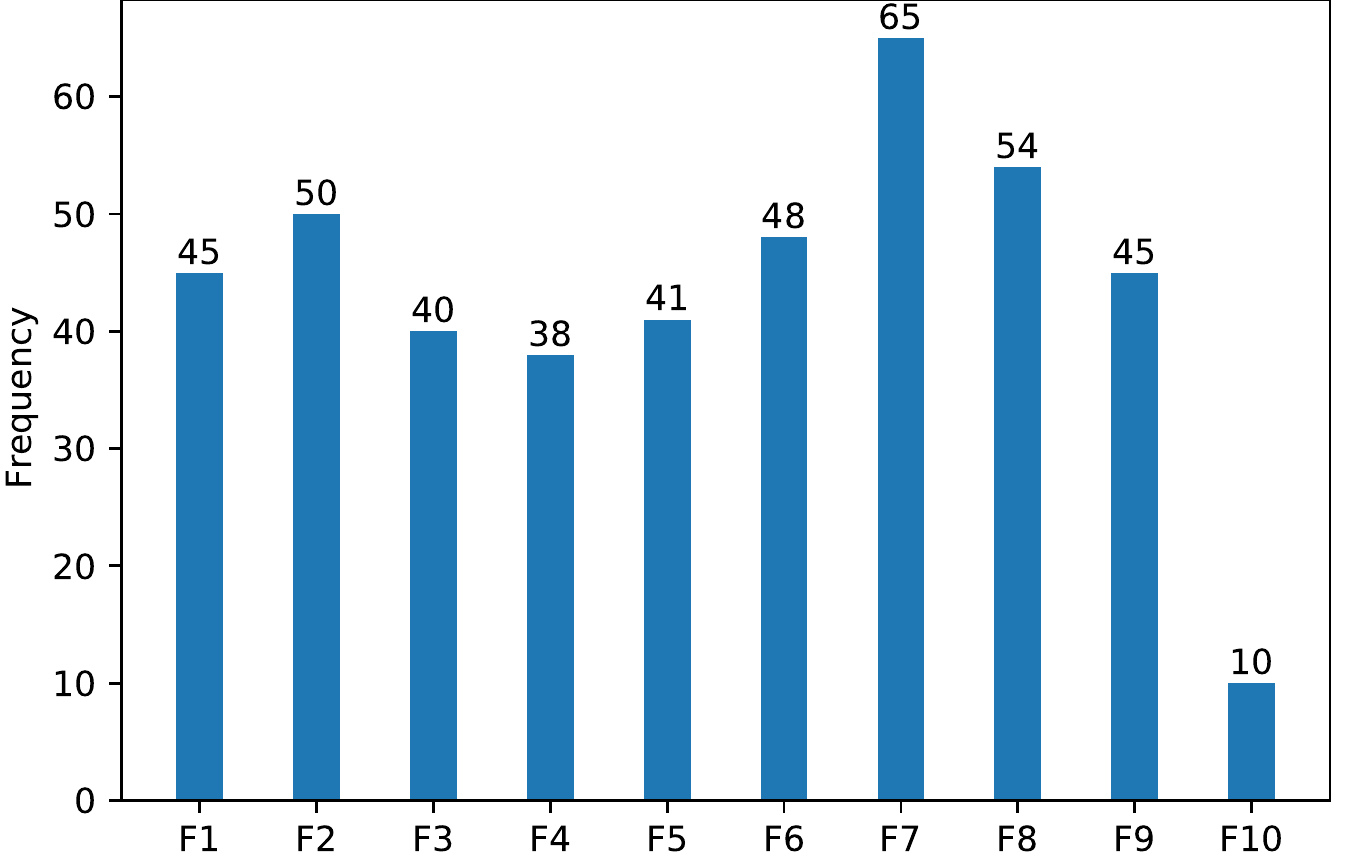}
    \caption{Functions participants with MCI or ADRD hoped the robot would help with. F1 = Help call family and friends; F2 = Help contact medical services; F3 = Help do shopping; F4 = Help order food; F5 = Help prevent falling; F6 = Remind to turn off stove top; F7 = Remind to take medicine; F8 = Emergency call service; F9 = Keep me company; F10 = Other.}
    \label{fig:bar_RobotFunctionHopedByPatients}
\end{figure}

Concerning the functions that people with MCI or ADRD hoped the robot could have, participants’ response are presented in Figure \ref{fig:bar_RobotFunctionHopedByPatients}. 
\begin{figure*}[t]
    \centering
    \includegraphics[width=11cm]{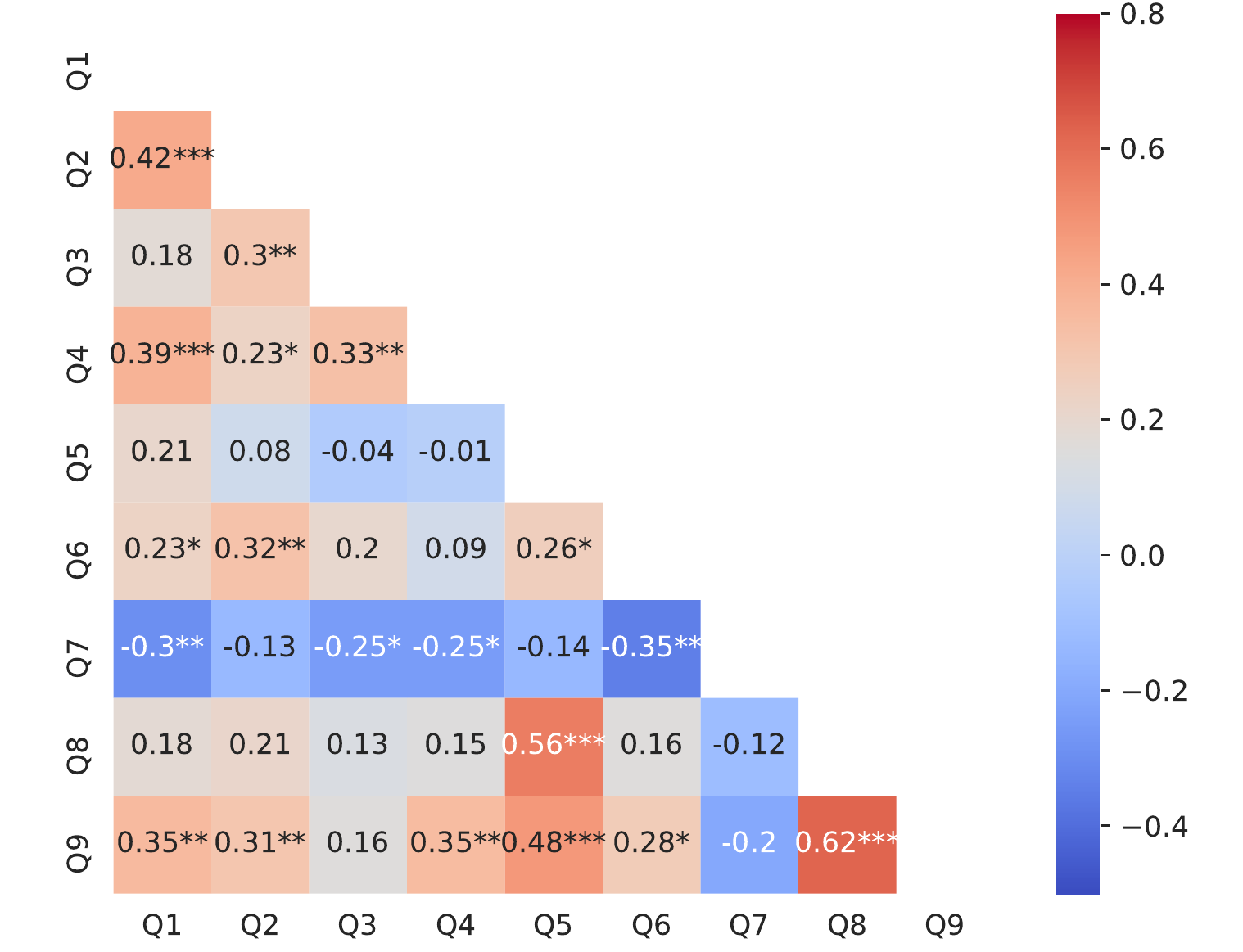}
    \caption{Correlation map among responses from people with MCI or ADRD. Please See Supplementary Appendix 1 for Items Q1-Q9. Asterisks denote statistically significant relationships between responses to two items (* at $0.05$ level, ** at $0.01$ level, *** at $0.001$ level).}
    \label{fig:correlationMap}
\end{figure*}
The top three desired functions in robots were reminders to take medicine, emergency call services, and helping contact medical services. All listed functions, except helping to order food, were desired by more than $50\%$ of people with MCI or ADRD. Additionally, in the "other" option, two participants suggested brain games, two suggested the robot helping with or encouraging to do physical exercise, two suggested the robot helping find continually misplaced items (e.g., glasses, keys, and TV remote), and three suggested the robot offering reminders to pay bills, appointments (e.g., appointment with doctor), the time (e.g., the day of the week and the year), to complete work activities, or how to do something. Other suggested functions included doing laundry, being someone to talk to, explaining why the person entered a third room, bringing things, naming their children to them, warning them if they were experiencing sleep apnea, monitoring for smoke/fire and leading to an exit, video search engine, and interacting with the television.
\\
With respect to ways to control the robot (tablet, speech, or both tablet and speech), there was $1$ participant who skipped this question. The majority ($50$ out of $80$) of people with MCI or ADRD would like to control the robot via both tablet and speech, while $29$ patients would prefer speech. No participants chose a tablet to control the robot.
\\
The Spearman’s correlations between responses to items Q1–Q9 are illustrated in Figure \ref{fig:correlationMap}. There was a statistically significant positive relationship between responses to item Q9 (i.e., "Would you like to follow the advice provided by a robot, such as the robot in Tammy in the video?") and responses to other items such as Q1, Q2, Q4, Q5, and Q8. There was a statistically significant negative relationship between responses to item Q7 and responses to items Q1, Q3, Q4, and Q6.
\\
\subsection{Perception of robot among caregivers and the general public}
The descriptive statistics for the responses of caregivers and the general public are reported in Table \ref{Table:PercentFreq_Caregiver_Others}. For both caregivers and the general public, the mean values of their ratings to all the items are greater than 3. Concerning the robot’s features, capabilities, and functions (i.e., items in Section B in Table \ref{Table:PercentFreq_Caregiver_Others}) to care for people with MCI or ADRD, most mean values of their ratings were greater than 4. Caregivers gave an overall higher rating than the general public for items B1d ($p=0.0441$), B1e ($p=0.0429$), B1f ($p=0.0145$), B2e ($p=0.0191$), B2g ($p=0.0005$), B2h ($p=0.0218$) and B3e ($p=0.0073$). With respect to the potential usefulness of robots in caring for people with MCI or ADRD, as shown in the rows associated with Items C1a–C1h of Table \ref{Table:PercentFreq_Caregiver_Others}, over half of both caregiver and general public participants thought the robot could be extremely useful to ADRD care. Moreover, caregivers overall rated in item C1f (i.e., "Detect when a person is becoming more lonely or isolated"; $p=0.0091$) higher than the general public.

{\onecolumn
\footnotesize{
\renewcommand{\arraystretch}{1.12}
\begin{longtable}[H]{p{0.6cm}p{1.8cm}p{1.8cm}p{1.8cm}p{1.8cm}p{1.8cm}p{1.5cm}}
\caption{Percentage of responses by caregivers and general public.} 
\label{Table:PercentFreq_Caregiver_Others} \\
\hline\noalign{\smallskip} {Items} & 1 & 2 & {3} & {4} & {5} & {Mean(SD)}\\ 
\noalign{\smallskip}\hline\noalign{\smallskip}
\endfirsthead
\multicolumn{7}{l}%
{\tablename\ \thetable{} -- continued from previous page} \\
\hline\noalign{\smallskip} Items & 1 & 2 & {3} & {4} & {5} & {Mean(SD)}\\
\endhead
\hline 
\multicolumn{7}{r}{{Continued on next page}} \\
\endfoot
\hline
\endlastfoot

\multicolumn{7}{l}{\textit{Section A: Acceptability}} \\
\hline
\multirow{2}{4em}{1} & $9$ $(3.67\%)$ & $56$ $(22.86\%)$ & $24$ $(9.80\%)$ & $109$ $(44.49\%)$ & $37$ $(15.10\%)$ & $3.46$ $(1.13)$ \\
& $17$ $(5.61\%)$ & $51$ $(16.83\%)$ & $29$ $(9.57\%)$ & $152$ $(50.17\%)$ & $35$ $(11.55\%)$ & $3.48$ $(1.10)$\\
\hline
\multirow{2}{4em}{2} & $25$ $(10.25\%)$ & $56$ $(22.95\%)$ & $38$ $(15.57\%)$ & $89$ $(36.48\%)$ & $29$ $(11.89\%)$ & $3.17$ $(1.22)$\\
& $27$ $(8.91\%)$ & $77$ $(25.41\%)$ & $35$ $(11.55\%)$ & $125$ $(41.25\%)$ & $30$ $(9.90\%)$ & $3.18$ $(1.20)$\\
\hline
\multirow{2}{4em}{3} & $6$ $(2.48\%)$ & $18$ $(7.44\%)$ & $45$ $(18.60\%)$ & $110$ $(45.45\%)$ & $54$ $(22.31\%)$ & $3.81$ $(0.97)$\\
& $9$ $(2.94\%)$ & $31$ $(10.13\%)$ & $54$ $(17.65\%)$ & $132$ $(43.14\%)$ & $67$ $(21.90\%)$ & $3.74$ $(1.02)$\\
\hline
\multirow{2}{4em}{4a} & $6$ $(2.45\%)$ & $18$ $(7.35\%)$&$9$ $(3.67\%)$&$106$ $(43.27\%)$&$103$ $(42.04\%)$& $4.17$ $(0.98)$\\
& $12$ $(3.92\%)$ & $10$ $(3.27\%)$ & $19$ $(6.21\%)$ & $117$ $(38.24\%)$ & $145$ $(47.39\%)$ & $4.23$ $(0.99)$\\
\hline
\multirow{2}{4em}{4b} & $9$ $(3.67\%)$ & $35$ $(14.29\%)$ & $26$ $(10.61\%)$ & $97$ $(39.59\%)$ & $78$ $(31.84\%)$ & $3.82$ $(1.14)$\\
& $13$ $(4.26\%)$ & $28$ $(9.18\%)$ & $31$ $(10.16\%)$ & $127$ $(41.64\%)$ & $97$ $(31.80\%)$ & $3.90$ $(1.10)$\\
\hline
\multirow{2}{4em}{4c} & $18$ $(7.41\%)$ & $50$ $(20.58\%)$&$37$ $(15.23\%)$&$76$ $(31.28\%)$ & $60$ $(24.69\%)$ & $3.46$ $(1.27)$\\
& $20$ $(6.58\%)$ & $48$ $(15.79\%)$ & $40$ $(13.16\%)$ & $103$ $(33.88\%)$ & $84$ $(27.63\%)$ &$3.62$ $(1.24)$ \\
\hline
\multirow{2}{4em}{4d} & $7$ $(2.87\%)$ & $11$ $(4.51\%)$ & $9$ $(3.69\%)$ & $95$ $(38.93\%)$ & $122$ $(50.00\%)$ & $4.29$ $(0.95)$\\
& $6$ $(1.97\%)$ & $19$ $(6.23\%)$ & $19$ $(6.23\%)$ & $112$ $(36.72\%)$ & $145$ $(47.54\%)$ & $4.23$ $(0.96)$\\
\hline
\multirow{2}{4em}{4e} & $8$ $(3.29\%)$ & $11$ $(4.53\%)$ & $12$ $(4.94\%)$ & $83$ $(34.16\%)$ & $125$ $(51.44\%)$ & $4.28$ $(0.99)$\\
& $9$ $(2.94\%)$ & $17$ $(5.56\%)$ & $21$ $(6.86\%)$ & $113$ $(36.93\%)$ & $141$ $(46.08\%)$ & $4.20$ $(1.00)$\\
\hline\noalign{\smallskip}
\multicolumn{7}{l}{\textit{Section B: Functionality}} \\
\noalign{\smallskip}\hline
\multirow{2}{4em}{1a} & $14$ $(5.71\%)$ & $18$ $(7.35\%)$ & $31$ $(12.65\%)$ & $81$ $(33.06\%)$ & $93$ $(37.96\%)$ & $3.93$ $(1.17)$\\
& $10$ $(3.27\%)$ & $25$ $(8.17\%)$ & $37$ $(12.09\%)$ & $93$ $(30.39\%)$ & $123$ $(40.20\%)$ & $4.02$ $(1.10)$\\
\hline
\multirow{2}{4em}{1b} & $7$ $(2.86\%)$ & $16$ $(6.53\%)$ & $25$ $(10.20\%)$ & $98$ $(40.00\%)$ & $90$ $(36.73\%)$ & $4.05$ $(1.01)$\\
& $4$ $(1.32\%)$ & $18$ $(5.92\%)$ & $27$ $(8.88\%)$ & $150$ $(49.34\%)$ & $89$ $(29.28\%)$ & $4.05$ $(0.88)$\\
\hline
\multirow{2}{4em}{1c} & $3$ $(1.22\%)$ & $13$ $(5.31\%)$ & $30$ $(12.24\%)$ & $101$ $(41.22\%)$ & $94$ $(38.37\%)$& $4.12$ $(0.91)$\\
& $9$ $(2.94\%)$ & $16$ $(5.23\%)$ & $36$ $(11.76\%)$ & $130$ $(42.48\%)$ & $99$ $(32.35\%)$ & $4.01$ $(0.98)$\\
\hline
\multirow{2}{4em}{1d *} & $3$ $(1.23\%)$ & $6$ $(2.46\%)$ & $29$ $(11.89\%)$ & $97$ $(39.75\%)$ & $106$ $(43.44\%)$ & $4.23$ $(0.85)$\\
& $5$ $(1.64\%)$ & $23$ $(7.54\%)$ & $27$ $(8.85\%)$ & $134$ $(43.93\%)$ & $105$ $(34.43\%)$ & $4.06$ $(0.96)$\\
\hline
\multirow{2}{4em}{1e *} & $2$ $(0.82\%)$ & $7$ $(2.86\%)$ & $12$ $(4.90\%)$ & $57$ $(23.27\%)$ & $164$ $(66.94\%)$ & $4.55$ $(0.79)$\\
& $4$ $(1.31\%)$ & $10$ $(3.28\%)$ & $15$ $(4.92\%)$ & $94$ $(30.82\%)$ & $175$ $(57.38\%)$ & $4.43$ $(0.84)$\\
\hline
\multirow{2}{4em}{1f *} & $4$ $(1.64\%)$ & $4$ $(1.64\%)$ & $9$ $(3.69\%)$ & $62$ $(25.41\%)$ & $164$ $(67.21\%)$ & $4.56$ $(0.79)$\\
& $3$ $(0.98\%)$ & $10$ $(3.27\%)$ & $20$ $(6.54\%)$ & $94$ $(30.72\%)$ & $172$ $(56.21\%)$ & $4.41$ $(0.84)$\\
\hline
\multirow{2}{4em}{1g} & $0$ & $4$ $(1.65\%)$ & $17$ $(7.00\%)$ & $61$ $(25.10\%)$ & $149$ $(61.32\%)$ & $4.54$ $(0.71)$\\
& $4$ $(1.31\%)$ & $7$ $(2.29\%)$ & $20$ $(6.54\%)$ & $76$ $(24.84\%)$ & $182$ $(59.48\%)$ & $4.47$ $(0.84)$\\
\hline
\multirow{2}{4em}{2a} & $5$ $(2.06\%)$ & $4$ $(1.65\%)$ & $8$ $(3.29\%)$ & $51$ $(20.99\%)$ & $171$ $(70.37\%)$ & $4.59$ $(0.81)$\\
& $5$ $(1.63\%)$ & $2$ $(0.65\%)$ & $12$ $(3.92\%)$ & $71$ $(23.20\%)$ & $212$ $(69.28\%)$ & $4.60$ $(0.75)$\\
\hline
\multirow{2}{4em}{2b} & $5$ $(2.05\%)$ & $2$ $(0.82\%)$ & $4$ $(1.64\%)$ & $43$ $(17.62\%)$ & $186$ $(76.23\%)$ & $4.68$ $(0.74)$\\
& $4$ $(1.31\%)$ & $3$ $(0.98\%)$ & $4$ $(1.31\%)$ & $54$ $(17.70\%)$ & $235$ $(77.05\%)$ & $4.71$ $(0.67)$\\
\hline
\multirow{2}{4em}{2c} & $1$ $(0.41\%)$ & $9$ $(3.70\%)$ & $10$ $(4.12\%)$ & $69$ $(28.40\%)$ & $146$ $(60.08\%)$ & $4.49$ $(0.79)$\\
& $4$ $(1.31\%)$ & $5$ $(1.64\%)$ & $8$ $(2.62\%)$ & $88$ $(28.85\%)$ & $190$ $(62.30\%)$ & $4.54$ $(0.75)$\\
\hline
\multirow{2}{4em}{2d} & $2$ $(0.82\%)$ & $3$ $(1.23\%)$ & $9$ $(3.69\%)$ & $67$ $(27.46\%)$ & $162$ $(66.39\%)$ & $4.58$ $(0.70)$\\
& $3$ $(0.99\%)$ & $3$ $(0.99\%)$ & $12$ $(3.95\%)$ & $95$ $(31.25\%)$ & $181$ $(59.54\%)$ & $4.52$ $(0.72)$\\
\hline
\multirow{2}{4em}{2e *} & $3$ $(1.23\%)$ & $2$ $(0.82\%)$ & $3$ $(1.23\%)$ & $40$ $(16.39\%)$ & $196$ $(80.33\%)$ & $4.74$ $(0.65)$\\
& $4$ $(1.31\%)$ & $2$ $(0.65\%)$ & $10$ $(3.27\%)$ & $69$ $(22.55\%)$ & $215$ $(70.26\%)$ & $4.63$ $(0.71)$\\
\hline
\multirow{2}{4em}{2f} & $1$ $(0.41\%)$ & $7$ $(2.87\%)$ & $18$ $(7.38\%)$ & $63$ $(25.82\%)$ & $153$ $(62.70\%)$ & $4.49$ $(0.79)$\\
& $3$ $(0.98\%)$ & $8$ $(2.61\%)$ & $22$ $(7.19\%)$ & $96$ $(31.37\%)$ & $167$ $(54.58\%)$ & $4.41$ $(0.82)$\\
\hline
\multirow{2}{4em}{2g ***} & $1$ $(0.41\%)$ & $7$ $(2.87\%)$ & $7$ $(2.87\%)$ & $55$ $(22.54\%)$ & $172$ $(70.49\%)$ & $4.61$ $(0.72)$\\
& $2$ $(0.66\%)$ & $7$ $(2.30\%)$ & $12$ $(3.93\%)$ & $112$ $(36.72\%)$ & $166$ $(54.43\%)$ & $4.45$ $(0.74)$\\
\hline
\multirow{2}{4em}{2h *} & $3$ $(1.23\%)$ & $2$ $(0.82\%)$ & $4$ $(1.64\%)$ & $27$ $(11.07\%)$ & $208$ $(85.25\%)$ & $4.78$ $(0.63)$\\
& $4$ $(1.31\%)$ & $3$ $(0.98\%)$ & $7$ $(2.29\%)$ & $54$ $(17.65\%)$ & $231$ $(75.49\%)$ & $4.69$ $(0.70)$\\
\hline
\multirow{2}{4em}{2i} & $1$ $(0.41\%)$ & $9$ $(3.69\%)$ & $15$ $(6.15\%)$ & $66$ $(27.05\%)$ & $149$ $(61.07\%)$ & $4.47$ $(0.81)$\\
& $2$ $(0.66\%)$ & $7$ $(2.30\%)$ & $17$ $(5.59\%)$ & $100$ $(32.89\%)$ & $168$ $(55.26\%)$ & $4.45$ $(0.77)$\\
\hline
\multirow{2}{4em}{2j} & $4$ $(1.64\%)$ & $5$ $(2.05\%)$ & $15$ $(6.15\%)$ & $54$ $(22.13\%)$ & $160$ $(65.57\%)$ & $4.52$ $(0.84)$\\
& $4$ $(1.32\%)$ & $7$ $(2.30\%)$ & $20$ $(6.58\%)$ & $85$ $(27.96\%)$ & $180$ $(59.21\%)$ & $4.45$ $(0.83)$\\
\hline
\multirow{2}{4em}{3a} & $4$ $(1.63\%)$ & $3$ $(1.22\%)$ & $8$ $(3.27\%)$ & $44$ $(17.96\%)$ & $183$ $(74.69\%)$ & $4.65$ $(0.75)$\\
& $3$ $(0.98\%)$ & $3$ $(0.98\%)$ & $10$ $(3.27\%)$ & $59$ $(19.28\%)$ & $225$ $(73.53\%)$ & $4.67$ $(0.69)$\\
\hline
\multirow{2}{4em}{3b} & $2$ $(0.82\%)$ & $4$ $(1.63\%)$ & $9$ $(3.67\%)$ & $54$ $(22.04\%)$ & $173$ $(70.61\%)$ & $4.62$ $(0.71)$\\
& $4$ $(1.31\%)$ & $1$ $(0.33\%)$ & $9$ $(2.95\%)$ & $62$ $(20.33\%)$ & $221$ $(72.46\%)$ & $4.67$ $(0.68)$\\
\hline
\multirow{2}{4em}{3c} & $2$ $(0.82\%)$ & $1$ $(0.41\%)$ & $5$ $(2.05\%)$ & $21$ $(8.61)$ & $215$ $(88.11\%)$ & $4.83$ $(0.55)$\\
& $4$ $(1.31\%)$ & $2$ $(0.66\%)$ & $3$ $(0.98\%)$ & $41$ $(13.44\%)$ & $249$ $(81.64\%)$ & $4.77$ $(0.63)$\\
\hline
\multirow{2}{4em}{3d} & $3$ $(1.22\%)$ & $4$ $(1.63\%)$ & $4$ $(1.63\%)$ & $47$ $(19.18\%)$ & $186$ $(75.92\%)$ & $4.68$ $(0.71)$\\
& $2$ $(0.65\%)$ & $2$ $(0.65\%)$ & $12$ $(3.92\%)$ & $68$ $(22.22\%)$ & $217$ $(70.92\%)$ & $4.65$ $(0.65)$\\
\hline
\multirow{2}{4em}{3e **} & $1$ $(0.41\%)$ & $8$ $(3.27\%)$ & $8$ $(3.27\%)$ & $82$ $(33.47\%)$ & $145$ $(59.18\%)$ & $4.48$ $(0.75)$\\
& $3$ $(0.98\%)$ & $5$ $(1.63\%)$ & $25$ $(8.17\%)$ & $122$ $(39.87\%)$ & $144$ $(47.06\%)$ & $4.33$ $(0.78)$\\
\hline
\multirow{2}{4em}{3f} & $3$ $(1.23\%)$ & $17$ $(7.00\%)$ & $32$ $(13.17\%)$ & $81$ $(33.33\%)$ & $95$ $(39.09\%)$ & $4.09$ $(0.99)$\\
& $4$ $(1.32\%)$ & $9$ $(2.96\%)$ & $34$ $(11.18\%)$ & $109$ $(35.86\%)$ & $112$ $(36.84\%)$ & $4.18$ $(0.89)$\\
\hline
\multirow{2}{4em}{3g} & $1$ $(0.41\%)$ & $6$ $(2.46\%)$ & $15$ $(6.15\%)$ & $77$ $(31.56\%)$ & $122$ $(50.00\%)$ & $4.42$ $(0.77)$\\
& $4$ $(1.31\%)$ & $1$ $(0.33\%)$ & $22$ $(7.21\%)$ & $107$ $(35.08\%)$ & $146$ $(47.87\%)$ & $4.39$ $(0.77)$\\
\hline
\multirow{2}{4em}{3h} & $2$ $(0.82\%)$ & $4$ $(1.65\%)$ & $10$ $(4.12\%)$ & $69$ $(28.40\%)$ & $153$ $(62.96\%)$ & $4.54$ $(0.73)$\\
& $5$ $(1.64\%)$ & $5$ $(1.64\%)$ & $12$ $(3.93\%)$ & $79$ $(25.90\%)$ & $192$ $(62.95\%)$ & $4.53$ $(0.80)$\\
\hline
\multirow{2}{4em}{3i} & $3$ $(1.24\%)$ & $6$ $(2.49\%)$ & $7$ $(2.90\%)$ & $67$ $(27.80\%)$ & $155$ $(64.32\%)$ & $4.53$ $(0.78)$\\
& $5$ $(1.64\%)$ & $2$ $(0.66\%)$ & $11$ $(3.61\%)$ & $69$ $(22.62\%)$ & $207$ $(67.87\%)$ & $4.60$ $(0.75)$\\
\hline
\multirow{2}{4em}{3j} & $3$ $(1.24\%)$ & $3$ $(1.24\%)$ & $13$ $(5.37\%)$ & $49$ $(20.25\%)$ & $166$ $(68.60\%)$ & $4.59$ $(0.77)$\\
& $5$ $(1.64\%)$ & $4$ $(1.31\%)$ & $15$ $(4.92\%)$ & $66$ $(21.64\%)$ & $202$ $(66.23\%)$ & $4.56$ $(0.80)$\\
\hline\noalign{\smallskip}
\multicolumn{7}{l}{\textit{Section C: Impact}} \\
\noalign{\smallskip}\hline
\multirow{2}{4em}{1a} & $0$ & $2$ $(0.82\%)$ & $5$ $(2.05\%)$ & $92$ $(37.70\%)$ & $143$ $(58.61\%)$ & $4.55$ $(0.58)$\\
& $2$ $(0.66\%)$ & $3$ $(0.99\%)$ & $14$ $(4.61\%)$ & $99$ $(32.57\%)$ & $178$ $(58.55\%)$ & $4.51$ $(0.70)$\\
\hline
\multirow{2}{4em}{1b} & $0$ & $4$ $(1.64\%)$ & $9$ $(3.69\%)$ & $87$ $(35.66\%)$ & $141$ $(57.79\%)$ & $4.51$ $(0.65)$\\
& $2$ $(0.66\%)$ & $2$ $(0.66\%)$ & $16$ $(5.26\%)$ & $100$ $(32.89\%)$ & $171$ $(56.25\%)$ & $4.50$ $(0.70)$\\
\hline
\multirow{2}{4em}{1c} & $1$ $(0.41\%)$ & $0$ & $7$ $(2.87\%)$ & $58$ $(23.77\%)$ & $176$ $(72.13\%)$ & $4.69$ $(0.57)$\\
& $1$ $(0.33\%)$ & $2$ $(0.66\%)$ & $8$ $(2.63\%)$ & $69$ $(22.70\%)$ & $217$ $(71.38\%)$ & $4.68$ $(0.59)$\\
\hline
\multirow{2}{4em}{1d} & $2$ $(0.82\%)$ & $0$ & $5$ $(2.06\%)$ & $20$ $(8.23\%)$ & $214$ $(88.07\%)$ & $4.84$ $(0.52)$\\
& $1$ $(0.33\%)$ & $3$ $(0.99\%)$ & $3$ $(0.99\%)$ & $35$ $(11.51\%)$ & $254$ $(83.55\%)$ & $4.82$ $(0.52)$\\
\hline
\multirow{2}{4em}{1e} & $1$ $(0.41\%)$ & $2$ $(0.82\%)$ & $9$ $(3.69\%)$ & $82$ $(33.61\%)$ & $148$ $(60.66\%)$ & $4.55$ $(0.65)$\\
& $2$ $(0.66\%)$ & $5$ $(1.64\%)$ & $9$ $(2.96\%)$ & $103$ $(33.88\%)$ & $174$ $(57.24\%)$ & $4.51$ $(0.71)$\\
\hline
\multirow{2}{4em}{1f **} & $4$ $(1.64\%)$ & $2$ $(0.82\%)$ & $10$ $(4.10\%)$ & $54$ $(22.13\%)$ & $168$ $(68.85\%)$ & $4.60$ $(0.76)$\\
& $3$ $(0.99\%)$ & $12$ $(3.96\%)$ & $16$ $(5.28\%)$ & $86$ $(28.38\%)$ & $175$ $(57.76\%)$ & $4.43$ $(0.85)$\\
\hline
\multirow{2}{4em}{1g} & $2$ $(0.82\%)$ & $3$ $(1.23\%)$ & $3$ $(1.23\%)$ & $52$ $(21.31\%)$ & $179$ $(73.36\%)$ & $4.69$ $(0.65)$\\
& $2$ $(0.66\%)$ & $6$ $(1.98\%)$ & $7$ $(2.31\%)$ & $69$ $(22.77\%)$ & $212$ $(69.97\%)$ & $4.63$ $(0.70)$\\
\hline
\multirow{2}{4em}{1h} & $0$ & $7$ $(2.88\%)$ & $16$ $(6.58\%)$ & $65$ $(26.75\%)$ & $153$ $(62.96\%)$ & $4.51$ $(0.75)$\\
& $3$ $(0.99\%)$ & $4$ $(1.32\%)$ & $12$ $(3.95\%)$ & $94$ $(30.92\%)$ & $186$ $(61.18\%)$ & $4.53$ $(0.73)$\\
\hline
\multirow{2}{4em}{2} & $15$ $(6.12\%)$ & $44$ $(17.96\%)$ & $64(26.12\%)$ & $69(28.16\%)$ & $45(18.37\%)$ & $3.36(1.17)$\\
& $22(7.24\%)$ & $74(24.34\%)$ & $75(24.67\%)$ & $74(24.34\%)$ & $46(15.13\%)$ & $3.16(1.19)$\\
\hline
\multirow{2}{4em}{3 *} & $7(2.88\%)$ & $55(22.63\%)$ & $56$ $(23.05\%)$ & $74$ $(30.45\%)$ & $34$ $(13.99\%)$ & $3.32$ $(1.09)$\\
& $21(6.98\%)$ & $74(24.58\%)$ & $65(21.59\%)$ & $80$ $(26.58\%)$ & $27$ $(8.97\%)$ & $3.07$ $(1.14)$\\
\hline
\multirow{2}{4em}{4 **} & $7$ $(2.86\%)$ & $29$ $(11.84\%)$ & $17$ $(6.94\%)$ & $52$ $(21.22\%)$ & $139$ $(56.73\%)$ & $4.18$ $(1.16)$\\
& $14$ $(4.62\%)$ & $37$ $(12.21\%)$ & $38$ $(12.54\%)$ & $80$ $(26.40\%)$ & $127$ $(41.91\%)$ & $3.91$ $(1.22)$\\
\bottomrule 
\end{longtable}
\begin{tablenotes}
\item \textit{Note:} Please see Supplementary Appendix 2 for Items A1-C4.
\item For the same item, the upper and lower row represent the responses from the caregiver and other public group, respectively.
\item For the last three items (i.e., C2, C3 and C4), 1=Strongly agree, 2=Somewhat agree,3=Neither agree nor disagree, 4=Somewhat disagree,5=Strongly disagree. Otherwise, 1=Extremely unlikely/ unimportant/ useless, 2=Somewhat unlikely/ unimportant/ useless, 3=Neither likely/ important/ useful nor unlikely/ unimportant/ useless, 4=Somewhat likely/ important/ useful, 5=Extremely likely/ important/ useful.
\item Asterisks denote statistically significant differences between rating responses of the caregiver group and the other public group (* at $0.05$ level, ** at $0.01$ level, *** at $0.001$ level.)
\end{tablenotes}
}
\twocolumn
}


\subsection{Ethical concerns}
Regarding the levels of agreement with the three statements related to ethical concerns (i.e., Items C2-C3 in Table \ref{Table:PercentFreq_Caregiver_Others}), participants’ responses were more scattered especially in terms of the first two statements, whereas the mean value of responses to these three questions were still greater than 3. Caregivers gave higher ratings overall than the general public group for items C3 ($p=0.0166$) and C4 ($p=0.0024$).

\subsection{Open-ended questions}
Additional comments, suggestions, and concerns that could be meaningful to the development of robot assistance for ADRD care were collected from caregivers and the general public. We received $111$ and $110$ comments from caregivers and the general public, respectively.

\subsubsection{Comments from caregivers}
Overall, caregivers supported the concept of developing a robot to assist in caring for people with MCI or ADRD and showed agreement in the potential application in the real world. A few people commented that they loved the idea that robots would be used to supplement human care interactions, not replace them, whereas there were several caregivers showing concern that family or other caregivers would be strongly tempted to rely too much on the robot for assistance. A few caregivers expressed that such robots would allow a bit more freedom to caregivers and also provide for the safety and well-being of individuals with cognitive impairment. One caregiver who has cared for many homebound elders with an inadequate support system mentioned that a well-designed robot could allow some seniors to stay at home and do it more safely. 

One dominant suggestion from the caregivers was to equip the robot with a menu of more human-like, soothing sounding voices instead of the current machine voice, which would be extremely uncomfortable to someone with ADRD. For example, one caregiver commented that, "Developing a more human sounding voice is needed. The machine/computer voice would be extremely startling to a patient." It was also suggested that the robot voice should be in regional language. Moreover, regarding the robot voice, one caregiver mentioned, "older people, with and without ADRD, tend to develop difficulties with hearing (especially for higher frequencies) and with rapidly processing of language. These people rely on ’lip reading’ to enhance cognition. A common request of older people ($<90$ years) is, ’Look at me and speak loudly and slowly.’ Pitch, volume, and cadence controls for the robot voice are each important. Of similar importance, but more challenging, is a display of realistic lip movement while speaking..." Caregivers also provided other suggestions such as the robot might be properly textured and have expressively mobile faces; the robot might be customized by changing its color or adding scent to the robot according to personal preference.

One big concern among caregivers was the cost of the robot. Some caregivers particularly mentioned that the robot for ADRD care should be affordable. One caregiver expressed, "To me, the underserved, under-cared for sector is the lower income sector. Those with little to no family. I believe they would benefit the most from this type of technology." Also, regarding cost, one caregiver participant asked if the robots are covered by insurance. Caregivers also showed other concerns and doubts such as the robot’s mobility on uneven surfaces or stairs, the size of the robot (e.g., "Not sure about size, not all patients have space for that size."), and the capability of the robot in understanding and conversing with someone with ADRD (e.g., "Such patients often use the wrong words for things, places and in an inconsistent manner. They use the wrong verbs and refer to people by incorrect names."). 
For better acceptability of the robot for ADRD care, caregivers commented that the robot should be introduced very slowly, carefully, and early to ensure that people using it understand it’s role and purpose. Several caregivers expressed concern that a robot would be a stranger and frightening to someone with ADRD. The robot should also be easy to use, friendly, and not creepy. For example, one comment of concern about the ease of use of the robot was as follows: "I can’t imagine an Alzheimer’s patient having anywhere near the cognitive or emotional ability to LEARN to use any of the functions, if learning such things is necessary." One caregiver also thought real-time technical support would be extremely important.

With respect to the capabilities and functions of robot for ADRD care, in addition to what was listed in the survey, caregivers suggested the robot could be used to help patients with staying hydrated, eating on schedule and with adequate nutrition, stress reduction (e.g., through meditation and yoga), and education regarding lifestyle changes (e.g., the importance of diet, water intake, exercise, meditation and sleep). It was also suggested it would be very meaningful if the robot could listen to patients’ repetitive life stories and respond, and provide alerts when someone had gotten up at night or was approaching the stairs.

\subsubsection{Comments from the general public}
Similar to caregivers, overall the general public showed positive attitudes towards the introduction of a robot for ADRD care, especially for cases where people with MCI or ADRD live alone, their caregivers are not currently available, or the robot can work as a companion. A few participants from the general public also suggested a more humanlike voice. One of the obvious differences in comments from the general public and caregivers is that a few participants were afraid of robots replacing human caregivers. For example, one participant commented that "...I am afraid too many families might place a relative with AD or dementia under the long term care of a robot...". They emphasized that human caregivers are always the most important. Another concern expressed by the general public was the cost of the robot. Participants mentioned that such robots should be available to all socio-economic levels. Associated with the concern of the cost of the robot, people in this group also questioned whether it is meaningful to use such a physically embodied robot (which could be expensive) in ADRD care, compared virtual agents such as Siri or Alexa. People in this group also shared with some suggestions for better acceptability. For example, the robot may be designed with different colors in terms of its physical appearance and two legs for being more human-like and mobile. There might be potential issues with charging the robot, the robot making a mistake, and patient privacy. Moreover, several participants suggested it would be important to acclimate the person to the robot as early as possible in the course of the disease to lessen the strangeness factor of having a robot around. For example, one participant commented, "Seems as if the robot would need to be in that person’s life at the very early stages of cognitive deterioration so as to allow comfort, trust $\&$ acceptance."

\section{Discussion}
\subsection{Robot acceptance}
People with MCI or ADRD showed an overall positive attitude towards the robot, Tammy, which is developed for ADRD care. As shown in Table \ref{Table:PercentFreq_Patient}, more than $50\%$ of people with MCI or ADRD gave a positive rating (i.e., Probably or Yes) to all the items except item Q7. Correspondingly, the mean values of their ratings to these items were greater than 3. For Item Q7 (i.e., "If you use the robot Tammy in the video, would you be afraid to break something of the robot?"), $53$ ($66.25\%$) out of 80 people with MCI or ADRD responded "No" or "Probably not," which indicates potentially low anxiety among these individuals in terms of using the robot \cite{heerink2010assessing}. As indicated in the Almere model \cite{heerink2010assessing}, a lower anxiety has a beneficial impact on the user’s positive attitude towards the robot and trust in the system. Noticeably, compared to the robot’s appearance (M=$4.56$), the robot’s voice (M=$3.61$) seemed to be less acceptable among people with MCI or ADRD. According to the comments about with robot’s voice provided by caregivers and the general public, we conjecture that the current machine voice in the robot Tammy is not comfortable to some individuals with MCI or ADRD.

Regarding the responses of people with MCI or ADRD to items Q1-Q9, there were statistically significant relationships between some responses as shown in Figure \ref{fig:correlationMap}. The responses to Item Q5 were statistically significantly positively related to responses to Item Q6 ($r=0.26$, $p<0.05$), Q8 ($r=0.56$, $p<0.001$), and Q9 ($r=0.48$, $p<0.001$), while the responses to Q8 and Q9 were statistically significant positively related as well ($0.62$, $p<0.001$). Questions Q5, Q6, Q8, and Q9 are closely associated with the construct intention to use (ITU), perceived ease of use (PEOU), social presence (SP) and perceived sociability (PS), and trust, respectively \cite{whelan2018factors}. Therefore, to increase the acceptability of using robots to assist ADRD care, robot developers should pay attention to the designs (e.g., user interface) that influence the user’s PEOU, SP, PS, and trust. Regarding Item Q6, responses from people with MCI or ADRD were statistically significantly positively related to responses to items Q1 and Q2 and statistically significantly negatively related to responses to item Q7. At the same time, responses from people with MCI or ADRD to Q9 were statistically significantly positively related to responses to items Q1, Q2, and Q4. Questions Q4 and Q7 are both closely associated with the construct anxiety, which is defined as evoking anxious or emotional reactions when using the robot \cite{whelan2018factors}. There is a statistically significant negative relationship between responses to Q4 and Q7. This makes sense considering that a high rating (e.g., 4=Not scary at all) to Q4 represents the robot is not scary at all, while a high rating (e.g., 5=Yes) to Q7 represents the user would be afraid to break something of the robot. The relationship between PEOU (Item Q6) and ITU (Item Q5), the relationship between PEOU (Item Q6) and anxiety (Q7), and the relationship between trust (Item Q9) and SP and PS(Item Q8) we found here agrees with the Almere model, a popular model developed to test the acceptance of assistive social agents by older adult users \cite{heerink2010assessing} in terms of the statistically significant positive/negative relationship.

With respect to caregivers and the general public, both groups showed an overall positive attitude towards using robots as a complement to ADRD care. Among the responses to the questions associated with robot’s current appearance (A1), voice (A2), and height (A3), as shown Table \ref{Table:PercentFreq_Caregiver_Others}, the mean ratings for A2 by both caregivers and the general public were relatively low compared to the mean ratings for A1 and A3. This is plausible considering the negative comments provided by both groups in the open-ended question. Regarding the likelihood of using the robot in the example activities (i.e., Items A4a–A4e in Table \ref{Table:PercentFreq_Caregiver_Others}), both mean ratings from both groups were greater than 4 (4=Somewhat likely) to all items except questions A4b and A4c. The low rating in A4b and A4c could be relevant to robot mobility issues, which was reported in the open-ended question by both groups. The high rating responses to Items C1a–C1h in Table \ref{Table:PercentFreq_Caregiver_Others} and the comments by caregivers in the open-ended questions confirmed the potential application of the robot for care of older adults with MCI or ADRD in terms of improving safety, providing companionship, and stimulation for cognitive and physical exercise, as well as reducing burden on the caregiver \cite{law2019developing}.
  

\subsection{User needs and requirements}
Regarding the primary user, people with MCI or ADRD, the top three required functions in robots are reminders to take medicine, emergency call service, and helping contact medical services as shown in Figure \ref{fig:bar_RobotFunctionHopedByPatients}. In addition to the functions listed in our survey, participants with MCI or ADRD specifically suggested other functions such as brain games, which was found to be the most popular feature among older adults with dementia \cite{cocsar2020enrichme}. Moreover, as indicated in the Subsection \ref{subsection_resultPerceptbyPatient}, 50 out of 79 participants with MCI or ADRD would like to control the robot through both tablet and speech, while the rest would like to control the robot through speech. It is a little outside of our expectations that none chose a tablet as the only means to control the robot. This encourages us to include the voice control together with the tablet control to use the robot. There was one participant with dementia who mentioned they were not sure if they had mild or moderate dementia. Considering one advantage of a robot caregiver is wide accessibility, we suggest robot developers take into consideration cognitive assessment as a function of the robot so that the user can use the robot to monitor disease progression.

With respect to the caregivers and the general public, among the list of features (i.e., Items B1a–B1g in Table \ref{Table:PercentFreq_Caregiver_Others}) that the robot has for ADRD care, {the features of the robot being able to be quiet (B1e), taking up no more room than a person while moving about (B1f), and being able to connect to internet (B1g) were rated as the top three highest} in terms of the mean value of response. With respect to the listed robot’s capabilities (i.e., Items B2a–B2j in Table \ref{Table:PercentFreq_Caregiver_Others}) to care for people with ADRD, over $60\%$ of caregivers rated as extremely important and over $50\%$ of the general public rated these as extremely important, with mean values all greater than $4.47$ and $4.41$, respectively. Thus, all these capabilities should be taken into consideration for implementation in the robot caregiver. In terms of the usefulness of listed functions (i.e., Items B3a–B3j in Table \ref{Table:PercentFreq_Caregiver_Others}) for a robot caring people with ADRD, both caregivers and the general public rated with a mean value greater than 4. Based on the mean value and response distribution of these items, the function of monitoring people’s medication to avoid errors in drug use (Item B3c) should be considered in the first position, which is consistent with the top desired functions by those with MCI or ADRD. The functions of monitoring people’s movements at home to reduce risks of falls (i.e., Item B3a) and monitoring the ambient environmental conditions (i.e., Item B3d) should also be put in high priority. Moreover, as reported in the rows of items C1a–C1h in Table \ref{Table:PercentFreq_Caregiver_Others}, approximately $60\%$ of caregivers and the general public thought the robot would be extremely useful for all eight items. Also, both caregivers and the general public provided the highest mean value, $4.84$ and $4.82$, respectively, for Item C1d (i.e., carry out emergency communication/alert message for people), which is consistent with the top desired functions by those with MCI or ADRD. Such requirements in robots of managing emergency situations has also been founded in previous relevant studies of the users’ needs and requirements \cite{faucounau2009caregivers,pigini2012service,manca2021impact,korchut2017challenges}. Over $93\%$ of caregivers rated the robot as somewhat or extremely useful to patients’ quality of life, quality and care, safety, and emergency communications, which has also been found in a multi-center study by Sancarlo et al. \cite{sancarlo2016mario}. Noticeably, caregivers provided a statistically higher rating response to Item C1f (i.e., Detect when a person is becoming more lonely or isolated) than the general public. We think this might be related to caregivers’ experiences of interacting with people with ADRD and their observations of the problem of loneliness among these individuals compared with the general public.

As we presented in the section of open-ended questions, both caregivers and the general public expressed concerns regarding the cost of the robot and that cognitive limitations due to dementia may inhibit the acceptability of the robot in ADRD care. These two factors were also found as the two biggest barriers to the use of assistive technology to support people with dementia in the literature review \cite{kruse2020evaluating}.

\subsection{Limitations and future work}
Some limitations in this survey study need to be taken into consideration in order to put the findings into perspective. {First, due to the online, anonymous survey, we cannot verify the identities of respondents, so there is no guarantee that they are people with ADRD and/or caregivers.} Second, it is possible that the responses in this study were biased. Because this survey was conducted online, the participants accessed were more likely familiar with technology (e.g., cellphone, computer, and ICT), which may influence users’ acceptance of a robot for dementia care. This selection bias may lead to the problem in acceptability work that the views of participants who find robots least acceptable may not be captured \cite{whelan2018factors}. 
Third, we used social media but by virtue of the architectural infrastructure the reach of a social media recruitment strategy was limited to a user’s social media community. For this reason, because the researchers reside in Tennessee (TN), USA, our social media recruitment yielded majority ($76.9\%$) responses from TN with fewer responses outside the TN community. Given $76.9\%$ of the respondents were from TN, there may be variations in acceptance and views across different states in US and different countries, and hence the sample representation may be skewed. 
Another limitation is that participants’ perceptions of the robot were merely based on watching a video instead of direct interaction with the robot. There might be some changes in their perceptions and attitudes towards robot after interacting with it. In the future, the evaluation of user experience (UX) among people with ADRD or MCI and their caregivers and health care providers will be a key step in the lifecycle of designing a robot for ADRD care \cite{hartson2018ux,khosla2017human}.

In the future, we will modify and develop a robot to supplement dementia care according to the user needs and requirements suggested by people with MCI or ADRD, caregivers, and the general public. For example, we will implement a menu of more human-like robot voices from which individuals can choose at the very beginning of interaction with the robot. In addition, both the caregivers and the general public expressed concerns that the robot would replace human caregiving and human companionship. One caregiver participant provided the following comment: "the video made me sad to think of older adults not having a caregiver to hold their hand." Therefore, we suggest providing the public, especially unpaid and paid caregivers, with sufficient education regarding the advantages of applying social robots to ADRD care in the future. It should be made clear that the introduction of a robot is aimed at supplementing their caregiving instead of totally replacing human caregivers, which is also indicated by the previous study among care personnel \cite{coco2018care}. For example, provided that the robot will be capable of performing routine tasks (e.g., displaying a picture/song during reminiscence therapy), the caregiver will be able to have more emotional, high-quality, personalized interaction with the person with MCI or ADRD, potentially decreasing caregiver burden. The successful application of social robots in ADRD care requires both the acceptance of people with ADRD and their caregivers.
\section{Conclusion}
In this work, we collected and compared the opinions held towards using robots to assist caring for people with ADRD among different groups of people (i.e., people with MCI or ADRD, caregivers, and the general public). All three groups showed overall positive attitudes towards the application of a robot in ADRD care. We also identified users’ needs and requirements indicated by people with MCI or ADRD, caregivers, and the general public. Additional comments, suggestions, and concerns from open-ended questions directed toward caregivers and the general public were also discussed. The results of users’ requirements and needs, as well as additional suggestions and concerns from the public, will be meaningful to the design and implementation of the social robot protocol for ADRD care.


%
%


\bibliographystyle{spbasic_unsorted}

\bibliography{reference}


\end{document}